\documentclass[sigconf]{acmart}

\usepackage{booktabs} % For formal tables
\usepackage{siunitx}
\newcommand{\argmax}{\operatornamewithlimits{arg\,max}}
\def\barr{\begin{tabular}{l}}
\def\earr{\end{tabular}}

\usepackage{subcaption}
\newcommand{\rowname}[1]% #1 = text
{\rotatebox{90}{\makebox[\tempdima][c]{\textbf{#1}}}}
\setcopyright{rightsretained}
\acmDOI{10.1145/nnnnnnn.nnnnnnn}

\acmISBN{978-x-xxxx-xxxx-x/YY/MM}

\acmConference[GECCO '20]{the Genetic and Evolutionary Computation Conference 2020}{July 8--12, 2020}{Under Review}
\acmYear{2020}
\copyrightyear{2020}

\acmPrice{15.00}

\begin{document}
\title{Divergent Search for Few-Shot Image Classification}
%Evolutionary Search for Problems and their Solutions}
%\titlenote{Produces the permission block, and
%  copyright information}
%\subtitle{Subtitle}
%\subtitlenote{The full version of the author's guide is available as
%  \texttt{acmart.pdf} document}

%%% The submitted version for review should be ANONYMOUS
%\author{Anonymous authors}
\author{Jeremy Tan}
%\authornote{Dr.~Trovato insisted his name be first.}
%\orcid{1234-5678-9012}
\affiliation{%
  \institution{Imperial College London}
%  \streetaddress{P.O. Box 1212}
%  \city{Dublin} 
%  \state{Ohio} 
%  \postcode{43017-6221}
}
\email{j.tan17@imperial.ac.uk}

\author{Bernhard Kainz}
%\authornote{The secretary disavows any knowledge of this author's actions.}
\affiliation{%
  \institution{Imperial College London}
%  \streetaddress{P.O. Box 1212}
%  \city{Dublin} 
%  \state{Ohio} 
%  \postcode{43017-6221}
}
%\email{webmaster@marysville-ohio.com}

%\

% The default list of authors is too long for headers.
%\renewcommand{\shortauthors}{B. Trovato et al.}

\begin{abstract}
When data is unlabelled and the target task is not known \textit{a priori}, divergent search offers a strategy for learning a wide range of skills. Having such a repertoire allows a system to adapt to new, unforeseen tasks. Unlabelled image data is plentiful, but it is not always known which features will be required for downstream tasks. We propose a method for divergent search in the few-shot image classification setting and evaluate with Omniglot and Mini-ImageNet. This high-dimensional behavior space includes all possible ways of partitioning the data. To manage divergent search in this space, we rely on a meta-learning framework to integrate useful features from diverse tasks into a single model. The final layer of this model is used as an index into the `archive' of all past behaviors. We search for regions in the behavior space that the current archive cannot reach. As expected, divergent search is outperformed by models with a strong bias toward the evaluation tasks. But it is able to match and sometimes exceed the performance of models that have a weak bias toward the target task or none at all. This demonstrates that divergent search is a viable approach, even in high-dimensional behavior spaces.   
\end{abstract}

%\keywords{ACM proceedings, \LaTeX, text tagging}

\maketitle

\section{Introduction}
Unsupervised learning is attractive because it avoids the need for manual data labelling; but also because human-crafted objectives are often only proxies for desired behaviors. For image classification, self-supervision and clustering are some of the most popular approaches that do not depend on labels. 
%Unsupervised learning is attractive because it avoids manual data labelling which is required in supervised learning; but also because human-crafted objectives are often only a proxy for the desired behavior. For image classification (of unlabelled data), some of the most popular approaches are self-supervision and clustering.
A lesser known alternative is divergent discriminative feature accumulation (DDFA). It searches for features that partition the data in novel ways and accumulates these features in an archive~\cite{szerlip2015unsupervised}. The advantage of a divergent search is that it explores as many diverse partition behaviors as possible. Learning a wide range of behaviors helps prepare a system when the target task is not known \textit{a priori}. In image classification, a single image can be classified in different ways based on its contents. An image of a dog at a park could be classified based on the breed of the dog, the dog's action/pose, the trees in the background, the season, the weather, or whether it is day or night time. Convergent algorithms generally settle on one partition of the set and may fail to consider other possible arrangements. In many cases, image-space similarities dominate and more specific/subtle features are hard to discover without explicit supervision. We aim to use divergent search to continually seek novel partitions rather than converging to the cardinal dimensions of variation.

%\begin{figure}
%    \includegraphics[width=\linewidth]{figures/nonDominantFeaturesCrop.pdf}
%	\caption{Examples from Imagenet\cite{russakovsky2015imagenet}. Each row of images contains a common feature, given in the class name. The left-most image in each row is entirely focused on the subject; whereas the rest of the images include extraneous, and sometimes more prominent, competing features. It can be difficult to learn the commonality in the latter group without explicit supervision.}
%	\label{figures:miniimagenet_examples}		
%\end{figure}

\begin{figure}
    \includegraphics[trim={3.5cm 14.5cm 7cm 2cm}, clip,width=\linewidth]{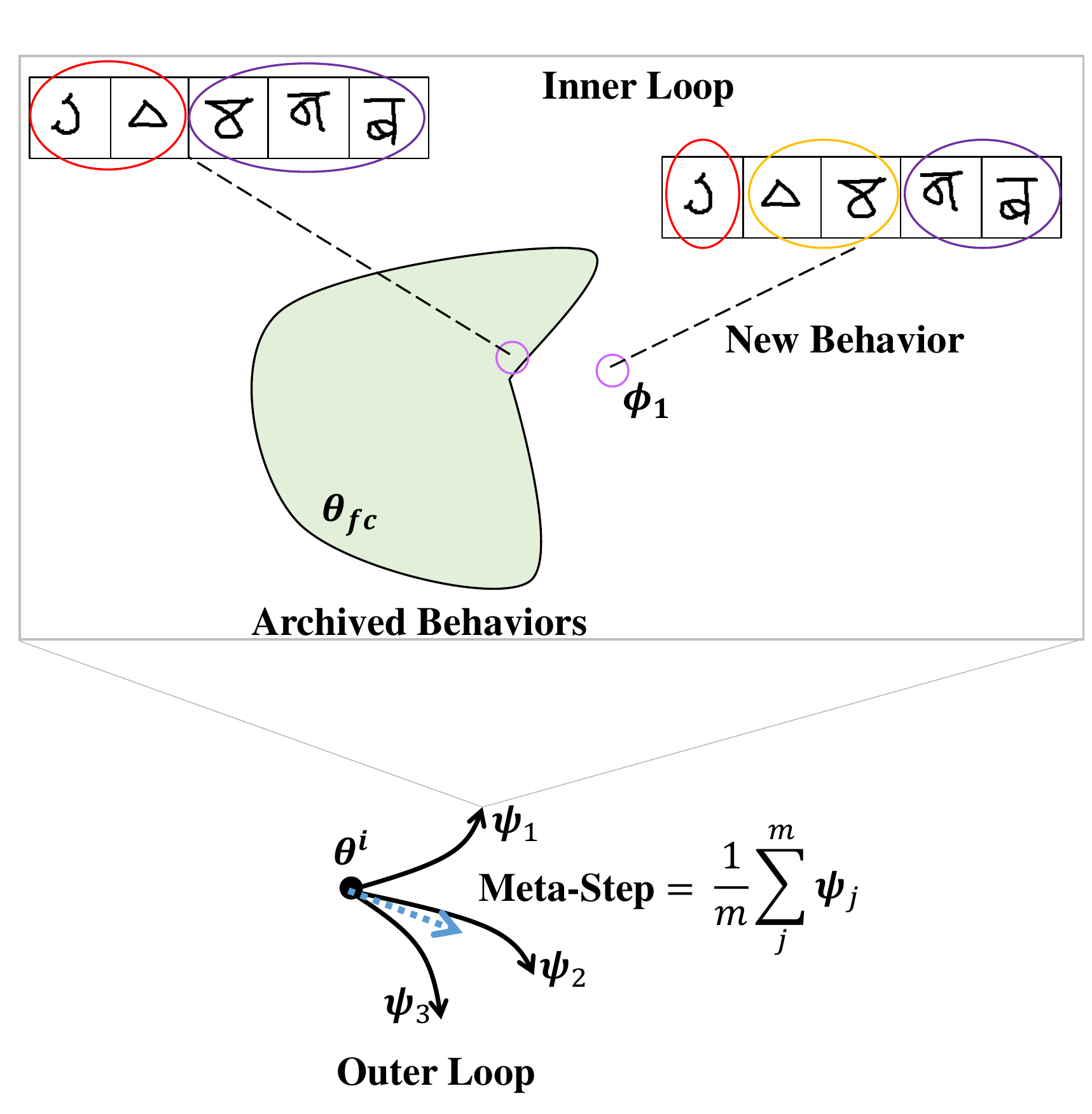}
	\caption{Overall framework with examples of Omniglot images. Each inner loop starts with the same model parameters, $\theta^i$. Then a search for novel behaviors begins: a fully connected layer, $\theta_{fc}$, acts as an index to all of the features stored in $\theta^i$, which acts as an archive of past behaviors. Another model, $\phi_1$, aims to find the parameterization of a behavior that is outside of the current capabilities of $\theta^i$. Then a learner $\psi_1$ is trained on that behavior. Finally the average of the gradients from each learner is used to update $\theta^i$ and the process repeats with an updated archive of behaviors.}
	\label{figures:framework}		
\end{figure}

Thus far, DDFA has only been demonstrated in very restricted settings (i.e. single layer features). One challenge of using divergent search with deep networks is the feature accumulation aspect. It is impractical to store copies of every network that produces a new partition behavior. It is even less feasible to search across a collection of networks for useful features to combine in the fine-tuning process. Recent work in meta-learning offers a simple way to learn from different tasks within a single network  \cite{finn2017model,nichol2018first}. These approaches aim to learn a set of features that is generally useful (or quickly adaptable) for a range of tasks, which may have never been presented during training. This lends itself well to divergent search. %integrating features learned from diverse tasks into one network.

Another challenge of using DDFA with deeper networks is the divergent search itself. Searching solely for novel partitions may suffice when the network's discriminative ability is inherently limited (i.e. single-layer features), but deeper networks are able to learn parameter settings that perfectly match random labels~\cite{zhang2016understanding}. To prevent accumulation of arbitrary and low utility features, a measure of quality must be included in the divergent search. In the absence of ground truth labels, robust consistency between network outputs has become a popular surrogate for quality~\cite{laine2016temporal,han2018co}.  %\textit{consistency} between networks has become a popular surrogate for quality in teacher-student models~\cite{laine2016temporal}. 
We jointly search for novelty and quality by optimizing a teacher network to exhibit behaviors that  
\begin{enumerate}
    \item have robust consistency with a student model, but
    \item cannot be learned by an archive of past behaviors (i.e. by reusing existing features)
\end{enumerate}
%We use these two objectives to search for novel, high-quality partition behaviors. 
Every novel behavior that is found in this search is integrated into the network. This increases its repertoire of discriminative behaviors and redefines novelty - pushing the search to new behaviors.

\section{Related Work}
\label{sect:relwork}
This work aims to translate the motivation from DDFA into more modern meta-learning frameworks. Both approaches are outlined  below.

\subsection*{Divergent Search}
One of the seminal works in this area is novelty search~\cite{lehman2011abandoning}, which eschews performance-based objectives in favor of finding novel behaviors. Since the proposal of novelty search, divergent strategies have become a central component of many evolutionary methods, particularly in the field of quality diversity (QD). QD argues that the strength of evolutionary computation lies in discovering new behaviors rather than optimizing toward a single targeted objective~\cite{pugh2016quality}. %QD argues that evolutionary computation is better at discovering new behaviors than optimizing toward a single targeted objective \ref{Pugh}. %QD algorithms have changed the way evolutionary computation is viewed; evolutionary search may be inefficient for optimizing toward a single targeted objective, but its greatest strength may lie in discovering new behaviors \ref{Pugh}. 
This paradigm has been used in many agent-based problem settings. For example, in robotics it has been used to learn a repertoire of behaviors spanning a robot's capabilities. This endows the robot with a range of pre-learned skills making it better prepared for uncertain terrain or new, unseen tasks \cite{cully2015robots}. Note the close parallels with unsupervised learning and meta-learning. 

Szerlip et al. proposed DDFA to harness divergent search for unsupervised image classification~\cite{szerlip2015unsupervised}. DDFA sets out to learn a repertoire of discriminative behaviors in hopes that some of these features will be useful for the target task. In their framework, each candidate feature is a single-layer feature (no hidden nodes) which outputs a scalar response for a given image. Behavior is measured as a feature's response to every image in the training set, forming a vector of length equal to the number of images. Novelty is then quantified as the distance between behavior vectors. In this way, new features are accumulated in an archive if their behavior vector is sufficiently different from existing features. Finally, a classifier can be trained on top of these features using labelled examples from a target task. 

Single-layer features are not sufficient for most computer vision applications~\cite{krizhevsky2012imagenet}. Scaling up to deeper networks is challenging because it is impractical to accumulate entire networks as opposed to individual features. The fine-tuning process also becomes unwieldy if the classifier must search for and combine features from an archive of networks. Furthermore, since deep networks have more discriminative capability than single-layer features, the space of possible behaviors could include virtually all possible partitions of the data. Running novelty search in this behavior space could lead to accumulation of features with very little semantic quality or utility. %This means that the vast majority of this large behavior space has very little semantic quality. Using novelty search in this vast behavior space of mostly meaningless partitions could lead to accumulation of features that are not useful for any real-world tasks.       
%Also, measuring novelty by running inference on the entire training dataset is computationally expensive for large datasets.

\subsection*{Meta-Learning}
Recent work in meta-learning could help to resolve some of DDFA's issues, specifically the integration of features for different tasks into a single network. Two very similar methods, model-agnostic meta-learning (MAML)~\cite{finn2017model} and Reptile (a play on words)~\cite{nichol2018first}, aim to learn a good initialization point which can quickly adapt to new tasks. Both use an inner loop which learns from a single task, and an outer loop which aggregates information from multiple tasks. We choose Reptile as the basis for this work for its simplicity and computational efficiency. Reptile learns parameter settings for a task by taking several ordinary gradient steps. This repeats for every task included in the outer loop, always restarting from the same initialization point. In the meta step, the initialization point is moved toward the mean of all task-tuned parameter settings.

\subsection*{Unsupervised Meta-Learning}
While meta-learning approaches are typically used in supervised settings, they offer attractive properties for unsupervised learning. Of primary interest is their ability to learn and generalize from tasks that are \textit{related}, but not \textit{exactly} the same as the the target task. This gives us more lenience when generating tasks for training (in an unsupervised way). Hsu et al. have exploited this fact for the purpose of unsupervised meta-learning~\cite{hsu2018unsupervised}. Their approach is based on clustering to automatically generate tasks for unsupervised model-agnostic meta-learning (CACTUs-MAML). They use existing self-supervised and clustering based methods to generate \textit{mock} tasks for training the meta-learning framework (in their case MAML). Note that the meta-learning model is typically small to prevent overfitting on few-shot tasks; but the self-supervised/clustering model can be much larger, creating a distillation effect~\cite{hinton2015distilling} from a more sophisticated model. The first step is to train a self-supervised/clustering method. Then k-means clustering is used to partition the \textit{embeddings} of the entire training dataset, in the latent space of the self-supervised/clustering network. Using these clusters as labels, regular tasks can be created for training the meta-learning framework. %To increase the variety of mock tasks, they also perform multiple runs of this k-means clustering with random scaling of the latent space dimensions. However, they show that this is not strictly necessary and provides nominal performance gains, even when comparing 100 runs to 1~\cite{hsu2018unsupervised}. However, the rescaling of the latent space is random, as opposed to approaches that explicitly optimize for partition novelty. 
Hsu et al. show that i) different self-supervised/clustering methods work better for different datasets and ii) the meta-learning performance is directly related to the quality of the mock tasks, i.e. the self-supervised/clustering method~\cite{hsu2018unsupervised}. As such, this process requires three distinct phases (self-supervision/clustering, k-means labelling, and meta-learning) with the potential need for trial-and-error in the self-supervised/clustering step. This process can be quite time-consuming and resource intensive.%, and cumbersome. 
%The clustering network can be much deeper, leading to better representations which are essentially being distilled into the meta-learner
%But going to 5 layers and 64 channels on miniimagenet still improved the supervised method when training with 1k iterations.

To circumvent the arduous training of CACTUs-MAML, Khodadadeh et al. propose unsupervised meta-learning for few-shot image classification (UMTRA)~\cite{khodadadeh2019unsupervised}. Their method uses domain knowledge as a substitute for explicit labels to construct tasks that are similar to the supervised tasks, but without the need for expensive labelling. First UMTRA exploits knowledge of the data distribution. Given the distribution of most few-shot datasets -- these include many classes and only a small number of examples within each class -- the samples in a small batch of images are likely to come from unique classes. UMTRA begins by sampling a small number of images and assigns unique labels to each image under the assumption that they are from unique classes. Their calculations indicate that a sample of five images from Omniglot has a 99.2\% chance of containing five unique classes (and a chance of 85.2\% for Mini-ImageNet)~\cite{khodadadeh2019unsupervised}. Data distributions of both datasets are shown in Table~\ref{tab:Datasets}. The second source of domain knowledge is knowledge of identity preserving transformations. Khodadadeh et al. use data augmentation to artificially expand this small sample of images~\cite{khodadadeh2019unsupervised}. The best results are achieved when AutoAugment~\cite{cubuk2019autoaugment} is used, which optimizes an augmentation policy to increase accuracy on a validation set. This combination of i) statistically-favorable sampling and ii) artificial upsampling, aims to create tasks which are as similar as possible to supervised tasks.

State-of-the-art approaches use self-supervision, clustering, or domain knowledge to construct tasks that closely approximate supervised tasks. The goal of training on these tasks is to give a bias toward the type of behaviors that the model will be evaluated on. Training and testing on the same behavior is a cornerstone of machine learning. But recent works in evolutionary computation have asked the question of whether undirected search, which lacks any specific objective, can discover useful behaviors on its own. Many of these investigations co-evolve tasks and solutions together in an open-ended fashion. For example generating mazes and their solvers~\cite{brant2017minimal} or courses with new terrains and agents that traverse them~\cite{wang2019paired}. Similarly, we aim to learn teacher and student models that jointly explore image classification behaviors. This behavior space is very high-dimensional and lacks the interpretability enjoyed by many reinforcement learning problems (e.g. agent sensors/states which correspond to different actions, or environment parameterizations that correspond to different obstacles). Using divergent search in this type of behavior space provides insight into its suitability for such problems and the remaining unmet needs. 

%Additional related papers to potentially include:
%diversity is all you need \ref{eysenbach}
%evolvability evolutionary strategies \ref{gajewski}
%goal switching: if working on task a, but improvement in task b, let it keep working on task b..
%go explore - after a behavior is found, enter imitation learning phase, with robustness to stochasticity..
%open ended evolutionary algorithms
%moco: momentum contrast for unsupervised visual representation learning
%contrastive predictive coding

%many evolutionary computation models use only the genetic aspect, but do not consider the extended phenotype?

\section{Methods}
In this work, unsupervised learning is formulated as a divergent search for novel partition behaviors. A meta-learning approach inspired by Reptile~\cite{nichol2018first} forms the basis of our method. An inner loop is used to derive a single task, while an outer loop is used to aggregate learning from multiple tasks. The overall framework is depicted in Figure~\ref{figures:framework}. 

\subsection{Inner-loop}\label{sec:inner}
Each iteration of the inner loop is an attempt to find a novel partition behavior. Three neural networks are used for this process. The \textbf{archive} is a fixed model which is parameterized by $\theta^i$. $\theta^i$ is the initial starting point for all three networks. All models follow the same architecture, described later in Section~\ref{sec:architecture}. The final, fully connected, layer of the archive, $\theta_{fc}$, is mutable. This means that the contribution of different higher-level features to a given logit can be changed. In this way, the fully connected layer acts as an index to all previous behaviors. This also includes exaptations of these archived features. In other words, features used for past behaviors can be repurposed for direct use in new behaviors. This allows the archive to span an even larger region of behavior space which includes all permutations of previous partition behaviors. 

The \textbf{teacher} model aims to find regions of the behavior space outside of the subspace already spanned by the archive (green region in Figure~\ref{figures:framework}). It is parameterized by $\phi$. Unfortunately it is difficult to analytically determine the `span' of the archive in behavior space. It is also prohibitively expensive to sample every possible index, $\theta_{fc}$. As such, the archive index and teacher model are jointly optimized. The archive index is optimized to match the behavior of the teacher. Let $p_{\phi}(x)$ denote the teacher's prediction for image $x$ where %(Eqn.~\ref{equation:tprediction}) and 
\begin{equation}
%\mathcal{L}_{\textrm{competitor}} = KL(R(x;w)||C(x'_u;w)).
p_{\phi}(x) = \argmax f(x;\phi), 
\label{equation:tprediction}
\end{equation}

and let $f_{\theta}(x)$ denote the archive outputs: 
\begin{equation}
%\mathcal{L}_{\textrm{competitor}} = KL(R(x;w)||C(x'_u;w)).
f_{\theta}(x) = f(x;\theta^i,\theta_{fc}).
\label{equation:aoutput}
\end{equation}

%(Eqn.~\ref{equation:aoutput}). 
The archive index is then optimized to reduce the categorical cross-entropy between its outputs and the teacher's predictions: %(Eqn.~\ref{equation:archiveIndexLoss}).

\begin{equation}
%\mathcal{L}_{\textrm{competitor}} = KL(R(x;w)||C(x'_u;w)).
\min_{\theta_\textrm{fc}}\; -\sum_{c=1}^{N=C} p_{\phi} \left(x \right) \log f_{\theta} \left(x \right)
\label{equation:archiveIndexLoss}
\end{equation}

Meanwhile, the teacher is optimized according to
\begin{equation}
%\mathcal{L}_{\textrm{competitor}} = KL(R(x;w)||C(x'_u;w)).
\min_{\phi}\; -JS(f_{\theta}(x)||f_{\phi}(x)), 
\label{equation:teacherLoss}
\end{equation} 
%Eqn.~\ref{equation:teacherLoss}, 
which aims to maximize the Jensen-Shannon (JS) divergence between the outputs of the archive and the teacher. The teacher and archive index are updated iteratively for a limited number of steps. In theory, a longer optimization process may find a better teacher behavior. To reduce computation time we restrict the number of iterations to 20 for all experiments. This is roughly double the number of steps that a model is typically given in the Reptile framework to train on a labelled classification task. As such, this limit is sufficient for learning new behaviors. 
%As such, this is a reasonable limit that is i) sufficient to reach new behaviors and ii) within the realm of what the learner can reach. The learner is further described at the end of this section. 

Another consideration is the optimizer, which in this case is Adam~\cite{kingma2014adam}. Typically the momentum is disabled for the optimizer used in the Reptile framework because it leads to poorer performance~\cite{nichol2018first}. However, in our case we find it helpful to restore momentum for the teacher optimizer to avoid oscillations between different behaviors \textit{within} the archive's reach.

Also note that the teacher is given a head-start by starting at a random mutation of the archive parameters through  %Namely, $\phi^i = \theta^i + s\eta$. 
\begin{equation}
\phi^i = \theta^i + s\eta.
\label{equation:noise}
\end{equation} 
Here $\eta$ is the parameter space noise, sampled from a normal distribution $\mathcal{N}(0,\sigma)$. The standard deviation, $\sigma$, is set low and increased until the teacher behavior differs from the initial archive behavior by at least one prediction within a batch of images. 
%The standard deviation is set low (e.g. on the same order of magnitude as the learning rate) and increased until the teacher predictions have at least one difference from the archive predictions for a batch of images. 
Each parameter has a noise scaling factor $s$, which scales the noise to a reasonable range. This is estimated by `safe mutations' which considers the magnitude of the gradient of all outputs with respect to a given weight~\cite{lehman2018safe}. 
%The noise for each parameter is also scaled by $s$, the sensitivity of the model outputs to that particular weight. This is based on safe mutations\cite{lehman2018safe} which calculates the gradient of the outputs with respect to a given weight. 

At the end of this process, a \textbf{learner} model is trained according to %Eqn.~\ref{equation:learnerLoss},
\begin{equation}
\min_{\psi}\; -\sum_{c=1}^{N=C} p_{\phi} \left(x \right) \log f_{\psi} \left(x \right),
\label{equation:learnerLoss}
\end{equation}

which attempts to match the teacher behavior. The learner model starts with the same parameters as the archive, $\psi^i = \theta^i$. The optimization of the learner is exactly the same as the standard inner loop in the Reptile framework~\cite{nichol2018first}, but using the teacher predictions in place of ground truth labels.

\subsection{Outer-loop}
The standard outer loop for Reptile~\cite{nichol2018first} simply takes the final parameter values of the learner from each inner loop iteration and calculates the mean (Eqn~\ref{equation:learnerMean}). Then a meta-step is taken toward this mean, depicted by the blue arrow in Figure~\ref{figures:framework}.

\begin{equation}
\overline{\psi} = \frac{1}{n}\sum_{i=1}^{n} {\psi}_i. 
\label{equation:learnerMean}
\end{equation}

Using ground truth labels, it is natural to weight each task equally. In an unsupervised setting, however, some tasks may be higher quality than others. In fact, a critical step in QD is measuring the novelty and quality of a behavior to determine whether it should be kept or discarded. We assess novelty by measuring how accurately the archive is able to match the teacher behavior. Meanwhile quality is judged based on the robustness of the teacher behavior. To estimate robustness, we measure how accurately the teacher behavior can be matched when a model is trained on noisy examples of the teacher behavior. During the joint optimization of the archive index and teacher (described in Section~\ref{sec:inner}), we include another copy of the archive that is trained on noisy teacher predictions (Eqn.~\ref{equation:robustLoss}). This is similar to Eqn.~\ref{equation:archiveIndexLoss}, except that i) optimization is no longer restricted to the last layer, $\theta_{fc}$, and ii) the teacher predictions are affected by Bernoulli noise in the inputs to the fully connected layer $\phi_{fc}$, i.e. dropout~\cite{srivastava2014dropout}. Using dropout in the final layer of the teacher will alter predictions that are less robust, while preserving predictions that are more robust.

\begin{equation}
\begin{split}
\min_{\theta}\; -\sum_{c=1}^{N=C} p_{\phi} \left(x,n_{fc} \right) \log f_{\theta} \left(x \right) \;,\; \\
\textrm{where} \; n_{fc}\sim \textrm{Bernoulli}(0.5)
\end{split}
\label{equation:robustLoss}
\end{equation}

Accuracy for each copy of the archive is calculated according to Eqn.~\ref{equation:accuracy_metric}. The value of a behavior is then calculated as the difference in accuracy between the two archive copies which are optimized according to Eqn.~\ref{equation:archiveIndexLoss} and Eqn.~\ref{equation:robustLoss}, which represent scores in novelty and robustness respectively. $V_{\phi} = A_{\textrm{robustness}} - A_{\textrm{novelty}}$. Note that a lower $A_{\textrm{novelty}}$ corresponds to a more novel behavior because the archive index was not able to find a past behavior which accurately fit the teacher behavior.

\begin{equation}
A_{\theta} = \frac{1}{m}\sum_{j=1}^{m} p_{\theta} \left(x \right) == p_{\phi} \left(x \right).
\label{equation:accuracy_metric}
\end{equation}

If the behavior value is negative, the behavior is discarded and a new inner loop starts. The final set of positive values is normalized to have a unit sum. These values are then used to calculate a weighted average of the learners: %Once enough positive values are acquired (based on the meta-batch size used in the standard Reptile framework~\cite{nichol2018first}), the behavior values are normalized to have a sum of 1. These normalized values are then used to weight each learned behavior (Eqn~\ref{equation:psi_mean}). 

\begin{equation}
\overline{\psi} = \sum_{i=1}^{n} v_i{\psi}_i \;,\; \textrm{where} \sum_{i=1}^{n} v_i = 1, v_i > 0.
\label{equation:psi_mean}
\end{equation}

This \textit{weighted} meta-step has parallels with evolution strategies (ES)~\cite{salimans2017evolution}. In ES, samples in parameter space are evaluated using a potentially non-differentiable loss function. Then the update is found by averaging all parameter sets, weighted by their loss. In our case, the non-differentiable loss function aims to measure quality and diversity of behaviors, and the loss evaluation is a complete inner loop.

%Also note that classification accuracy provides a \textit{discrete} measurement of prediction consistency. This conflates behavior to discrete bins. In other words, different parameter settings that produce the same set of predictions are considered equivalent in behavior space. As such, the index is used to search the archive for the behavior bin which best matches the teacher behavior bin. 
%This reduction in behavior space is not as helpful as in common reinforcement learning tasks. As such, a learner must produce  This allows the archive to dominate within bins of behavior. For instance, even if the learner has a slightly lower loss than the archive, the archive will dominate (value = 0) unless the difference is significant enough to change an image prediction. This is similar in spirit to methods that support \textit{local} competition within a small bin of behavior~\cite{lehman2011evolving,mouret2015illuminating}.   
%accuracy conflates behavior in the same way as niche archive or discretization.

%Add Algorithm blocks for inner and outer loop
\subsection{Model Architecture}\label{sec:architecture}
\begin{figure}[htb]
    \includegraphics[trim={9cm 0cm 0cm 3cm},clip,width=\linewidth]{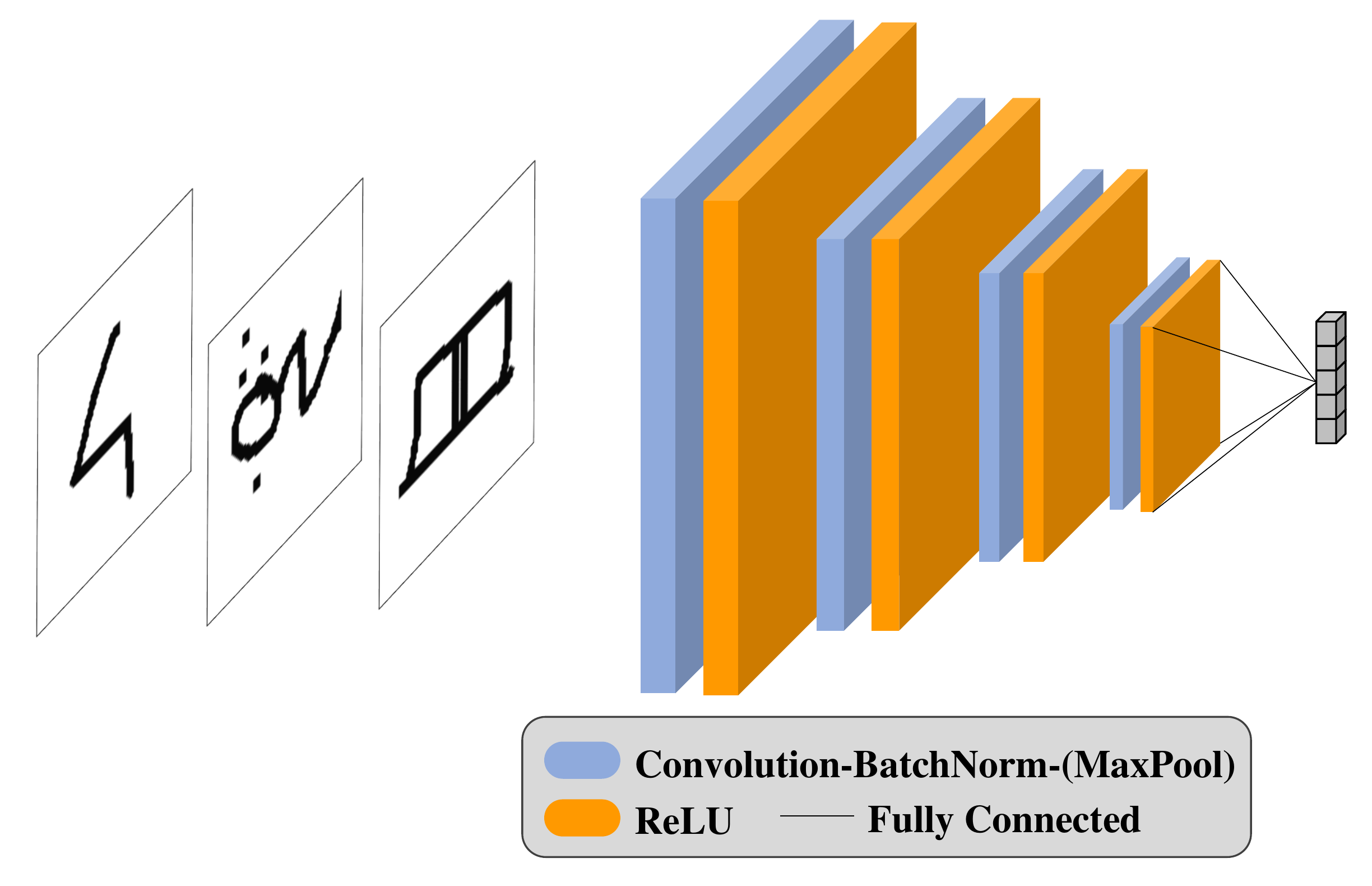}
	\caption{Network architecture with Omniglot example inputs.}
	\label{figures:network}		
\end{figure}

The models used in our experiments follow the same architecture and hyperparameter settings as the original Reptile framework~\cite{nichol2018first}. A neural network is used with four convolutional layers, followed by a fully connected layer (Figure~\ref{figures:network}). Each convolution is followed by batch normalization and a ReLU activation. In the case of Mini-ImageNet, there is also a max-pooling operation before the ReLU. The convolutional layers have 64 feature channels for the Omniglot dataset and 32 for Mini-ImageNet.

\section{EXPERIMENTAL STUDIES}

\subsection{Datasets} %TODO: very brief description of Mini-Imagenet and Omniglot. Avoid redundancy with in-text data description below. 
Two of the most commonly used datasets for few-shot learning are Omniglot~\cite{lake2015human} and Mini-ImageNet~\cite{vinyals2016matching,russakovsky2015imagenet}. Each of these datasets consists of many classes and relatively few examples (see Table~\ref{tab:Datasets}). Omniglot images are hand-written characters from many different alphabet systems, captured in grayscale with dimensions $28x28$ pixels. Mini-ImageNet consists of natural images of different subjects (e.g. golden retriever, poncho, school bus) captured in color with dimensions $84x84$ pixels. 

\begin{table}[htb]
    %\centering
    \caption{Dataset specifications.}
    \begin{tabular}{c|c|c} 
      %\textbf{Dataset} & \def\arraystretch{1} \barr \textbf{Pre-training}\\    \textbf{Classes} \earr & \def\arraystretch{1} \barr %\textbf{Evaluation}\\    \textbf{Classes} \earr & \def\arraystretch{1} \barr \textbf{Images per}\\    \textbf{Class} \earr\\
      \textbf{Dataset} & Omniglot & Mini-ImageNet \\
      \hline
      Pre-training Classes & 1200 & 64 \\ 
      Evaluation Classes & 423 & 36\\
      Images per Class & 20 & 600 \\
      %\hline

    \end{tabular}%
    \label{tab:Datasets}

\end{table}%

\subsection{Image Sampling}\label{sec:sampling}
\begin{figure}
    \includegraphics[width=\linewidth]{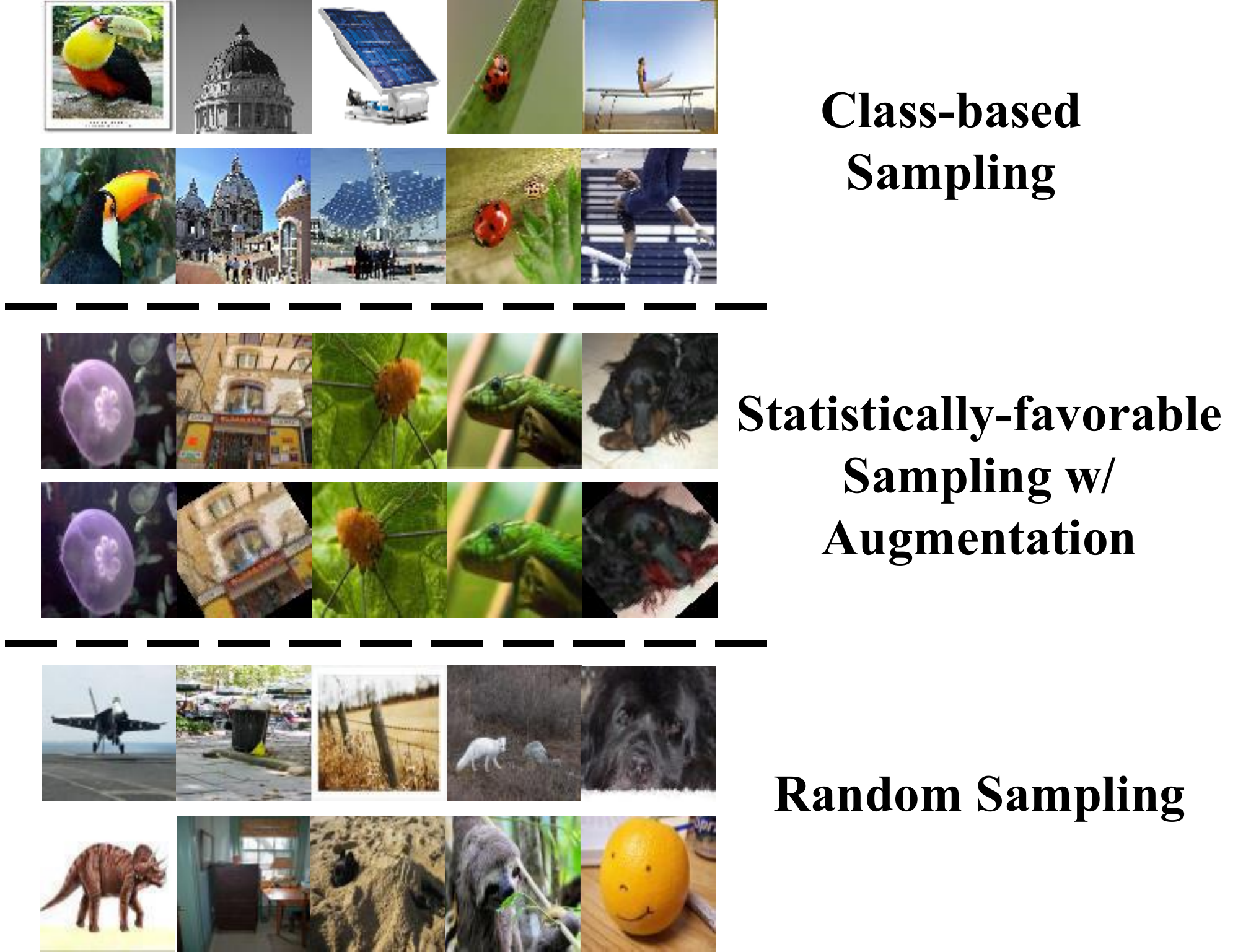}
	\caption{Examples from Mini-ImageNet based on different sampling schemes. Each row represents a batch.}
	\label{figures:samples}		
\end{figure}

An important consideration is how images are sampled for each inner loop. As mentioned in Section~\ref{sect:relwork}, UMTRA exploits knowledge of the data distribution to sample small batches which are likely to contain images from different classes. Figure~\ref{figures:samples} illustrates that UMTRA's statistically-favorable sampling, combined with augmentation, simulates genuine class-based sampling. Instead of relying on statistically-favorable sampling, we wish to simulate a case where sample size can no longer be exploited. %We examine their calculation to determine a sample size which is no longer favorable.%Instead of relying on knowledge of the data distribution, we aim to evaluate the proposed method in settings where statistically-favorable sampling cannot be exploited. As such, we use the same calculation from UMTRA to determine a sample size which no longer gives a favorable batch distribution. 
The probability that each sample in a batch comes from a different class is 
\begin{equation}
%V_C = \sum_{j=1}^{m} \argmax(R(x;\phi^f)) == \argmax(C(x;\theta^f))
P = \frac{c!\cdot m^N\cdot(c\cdot m - N)!}{(c-N)!\cdot (c\cdot m)!},
\label{equation:samplingProb}
\end{equation}

%given in Eqn.~\ref{equation:samplingProb}, 
where $c$ is the number of classes, $m$ is the number of images in each class, and $N$ is the sample size~\cite{khodadadeh2019unsupervised}.

By using a sample size of 20 images for Mini-ImageNet and 90 images for Omniglot, we can reduce the chance that all images are from unique classes to about 3\%. We use this setting, depicted in the last row of Figure~\ref{figures:samples}, for the proposed method. This distribution will likely contain multiple images from the same class, but it will also likely contain more classes than available logits. This means that pre-training tasks will almost certainly violate class-boundaries. These conditions are not favorable for inducing a bias toward class-oriented evaluation tasks. However, it simulates the most general case of unsupervised learning, where the number of classes and the distribution among classes is not known \textit{a priori}.

%a sample size of 20 has only a 3.6\% chance of containing all unique classes. To reach an equivalent probability (3.9\%) for Omniglot (1200 classes, 20 images/class) a sample size of 90 is required. As such, the proposed method uses a sample size of 20 for Mini-ImagNet and 91 for Omniglot. 

\subsection{Evaluation}
A few-shot learning task is used to evaluate whether the algorithm has learned useful features. In this setting, $N$ classes are selected and $K+1$ images are sampled from each class. The model is given the first $K$ images (from each class) along with their ground truth labels. After the model is fine-tuned using these labelled examples, the model must predict the correct class for the final ($+1$) image in each class. Note that `fine-tuning' will be used to refer to the learning that occurs during the evaluation stage (using the $K$ labelled examples); meanwhile `pre-training' will be used to refer to all learning that occurs before the evaluation stage. The reported accuracy metric is the average across 10,000 tasks, where each task uses a random set of $N$ classes. Note that the pre-training classes and test classes are disjoint sets.%In the standard supervised setting, the classes used in pre-training are separate from those used in evaluation. In the unsupervised case, ground truth labels are never used during pre-training. As such, evaluation on seen and unseen classes can both be informative.

\subsection{Benchmark Methods}
Most existing approaches, whether supervised or unsupervised, attempt to give the model a bias toward tasks that are similar to the evaluation tasks. In contrast, the proposed method is not biased toward any particular type of classification, but instead tries to learn as many discriminative behaviors as it can. A major question that arises is: \textit{whether the features learned by an undirected, divergent search are useful for downstream tasks designed by humans}. In particular, how does it compare with methods that induce a strong/weak bias toward the evaluation tasks or indeed away from the evaluation tasks. We use four main benchmarks to represent biases of varying strength: i) fully supervised, ii) UMTRA, iii) random initialization, and iv) random labels.

\subsubsection{Fully Supervised}
The fully supervised setting (Reptile~\cite{nichol2018first}) represents the most explicit way to create an inductive bias toward tasks that are similar to the evaluation tasks. Supervised pre-training uses labelled images to learn tasks that are essentially equivalent to the evaluation tasks, but using different classes. 

\subsubsection{Unsupervised meta-learning for few-shot image classification.} 
UMTRA uses domain knowledge as a substitute for explicit labels to construct tasks that are similar to the evaluation tasks. It combines a) statistically-favorable sampling and b) artificial upsampling through data augmentation~\cite{khodadadeh2019unsupervised}. We re-implement UMTRA within the Reptile~\cite{nichol2018first} framework for a more direct comparison. The original UMTRA uses AutoAugment~\cite{cubuk2019autoaugment} to create new realistic samples. AutoAugment has recently surpassed state-of-the-art scores on formidable challenges such as ImageNet~\cite{russakovsky2015imagenet}, purely through data augmentation. In our case, we restrict augmentations to a more modest set of standard transformations, including random combinations of:

\begin{itemize}
	\item Horizontal flipping
	\item Hue, saturation, brightness and contrast adjustment by a random factor within ranges [-0.08,0.08],[0.6,1.6],[-0.05,0.05], and [0.7, 1.3] respectively
	\item Random rotation by an angle within [$-\frac{\pi}{4}$, $\frac{\pi}{4}$]
	\item Random cropping ranging from 1\% to 20\%
\end{itemize}

While the original UMTRA uses strong augmentation to effectively `generate' additional examples of each class, our low-augmentation setting is more akin to teaching invariance to simple transformations. We use this baseline to represent a weaker bias toward evaluation tasks and include the original UMTRA results for reference. 

\subsubsection{Random Initialization} 
Random initialization represents an unbiased baseline, i.e. the performance of a model without pre-training. This method learns only from the fine-tuning during evaluation. 

\subsubsection{Fixed Random Labels} 
The fixed random labels approach represents a bias \textit{away} from the evaluation tasks. Assigning fixed random labels to the images creates new tasks that are very unlikely to overlap with the type of tasks used in evaluation. In these randomly labelled tasks, images with the same label are likely to come from different classes and the same class may appear under more than one label. As such, class-specific features are no longer reliable signals for discrimination. Instead, the model may rely on extraneous information in the images (e.g. noise or background objects). Note that, in some cases, pre-training on random labels (i.e. `memorizing') can actually lead to an increase in accuracy for a supervised task as compared with starting from a random initialization~\cite{pondenkandath2018leveraging}.

To summarize, (i) fully supervised and (ii) low-augmentation UMTRA represent strong and weak biases toward the evaluation tasks respectively; (iii) random initialization represents an unbiased starting point; and (iv) random labels acts as a bias away from the evaluation tasks. Comparing a divergent strategy to this spectrum of pre-training approaches allows us to assess how useful the discovered features are for real, human-designed tasks.

%novelty search with local competition needs to avoid global competition bc they use a global fitness function - but bc our fitness function is agreement, each individually naturally competes within its own niche? If a niche is more useful, it will defend its area more effectively, whereas if it is not effective it will be more prone to replacement by a new solution. 

%The resource, which is analogous to a generated task in the work by Hsu et al. The resource is optimized to exhibit a partition behavior that matches the organism, but escapes the capability of the competitor. The competitor and organism represent novelty and quality respectively.  
%The \textbf{competitor} tries to use existing features to match \textbf{resource} while the organism the the organism and the competitor start as identical networks, but the organism can freely change all parameters while the competitor can only adjust the last layer, making the best use of existing features; is a network that can only change its last layer to best use existing features  

\section{Results}
Evaluation was performed using $N=5$ classes and $K=5$ fine-tuning examples ($k=1$ was also tested). The results are summarized in Table~\ref{tab:OmniglotResults} and Table~\ref{tab:MiniImageNetResults}. The primary benchmark methods are shown in bold and below them are additional results summarized from literature~\cite{hsu2018unsupervised,khodadadeh2019unsupervised}. Note that bidirectional generative adversarial network (BiGAN)~\cite{donahue2016adversarial}, adversarially constrained autoencoder interpolation (ACAI)~\cite{berthelot2018understanding}, and DeepCluster~\cite{caron2018deep} are different self-supervised/clustering approaches. The representation learned from these methods can be used in several ways, e.g. a linear classifier trained on top of the representation or a k-nearest neighbors approach. Here, only the best and worst scores are included. For full details, refer to the original work by Hsu et al.~\cite{hsu2018unsupervised}. Note that despite the effort required to train these various methods, several of them produce scores below random initialization.

\begin{table}[htb]
\caption{Omniglot results for $N=5,K=5$ and $N=5,K=1$ evaluation tasks. Bottom results summarized from \cite{hsu2018unsupervised,khodadadeh2019unsupervised}.}
\begin{tabular}{lllll}
\textbf{Omniglot}\\ \hline
Algorithm (way-shot)            &(5,1)  &       & (5,5) &  \\ 
                                & train & test  & train & test \\\hline
\textbf{Fixed Random Labels}    & 21.11 & 20.84 & 26.05 & 25.60 \\
\textbf{Random Initialization}  & 34.11 & 33.64 & 66.28 & 64.29 \\
%trying to update proposed result
\textbf{\textit{Proposed}}      & 59.53 & 55.79 & 80.87 & 77.85 \\
\textbf{UMTRA-Reptile Low Aug}  & 66.21 & 65.26 & 85.32 & 83.48 \\
\textbf{Fully Supervised}       & 98.32 & 97.64 & 99.54 & 99.68 \\
\hline
Random Initialization MAML      & - -   & 52.50 & - -   & 74.78 \\
BiGAN worst scores              & - -   & 40.54 & - -   & 58.52 \\
BiGAN best scores               & - -   & 49.55 & - -   & 68.72 \\
BiGAN CACTUs-MAML               & - -   & 58.18 & - -   & 78.66 \\
ACAI worst scores               & - -   & 51.95 & - -   & 71.09 \\
ACAI best scores                & - -   & 61.08 & - -   & 81.82 \\
ACAI CACTUs-MAML                & - -   & 68.84 & - -   & 87.78 \\
UMTRA-MAML                      & - -   & 83.80 & - -   & 95.43 \\

%summarized version 
%Unsupervised worst scores       & - -   & 40.54 & - -   & 58.52 \\
%Unsupervised best scores        & - -   & 61.08 & - -   & 81.82 \\
%ACTUs-MAML best scores         & - -   & 68.84 & - -   & 87.78 \\

\hline

\end{tabular}
\label{tab:OmniglotResults}
\end{table}

%Additional results are summarized from related works~\cite{hsu2018unsupervised,khodadadeh2019unsupervised}.   

%Different strategies of fine-tuning on top of representations learned by unsupervised methods (e.g. BiGAN, DeepCluster) are .   

\begin{table}[]
\caption{Mini-ImageNet results for $N=5,K=5$ and $N=5,K=1$ evaluation tasks. Bottom results summarized from \cite{hsu2018unsupervised,khodadadeh2019unsupervised}.}
\begin{tabular}{lllll}
\textbf{Mini-ImageNet}\\ \hline
Algorithm (way-shot)            &(5,1)  &       & (5,5) &  \\ 
                                & train & test  & train & test \\\hline
\textbf{Fixed Random Labels}    & 25.78 & 24.46 & 36.97 & 32.49 \\
\textbf{Random Initialization}  & 26.02 & 24.59 & 39.93 & 36.52 \\
\textbf{\textit{Proposed}}      & 33.74 & 30.90 & 47.58 & 43.41 \\
\textbf{UMTRA-Reptile Low Aug}  & 34.13 & 31.45 & 46.84 & 42.15 \\
\textbf{Fully Supervised}       & 64.24 & 50.65 & 78.52 & 66.33 \\
\hline
Random Initialization MAML      & - -   & 27.59 & - -   & 38.48 \\
BiGAN worst scores              & - -   & 22.91 & - -   & 29.06 \\
BiGAN best scores               & - -   & 27.08 & - -   & 33.91 \\
BiGAN CACTUs-MAML               & - -   & 36.24 & - -   & 51.28 \\
DeepCluster worst scores        & - -   & 22.20 & - -   & 23.50 \\
DeepCluster best scores         & - -   & 29.44 & - -   & 42.25 \\
DeepCluster CACTUs-MAML         & - -   & 39.90 & - -   & 53.97 \\
UMTRA-MAML                      & - -   & 39.92 & - -   & 50.73 \\

%summarized version
%Unsupervised worst scores       & - -   & 22.20 & - -   & 23.50 \\
%Unsupervised best scores        & - -   & 29.44 & - -   & 42.25 \\
%CACTUs-MAML best scores         & - -   & 39.90 & - -   & 53.97 \\
\hline

\end{tabular}
\label{tab:MiniImageNetResults}
\end{table}

For both datasets, the proposed divergent search strategy achieves similar performance to methods with a weak bias toward evaluation tasks (low-augmentation UMTRA). However, methods with a stronger bias (fully supervised, original UMTRA, and CACTUs-MAML methods), still perform significantly better. Meanwhile the fixed random labels approach actually degrades performance compared to an unbiased start from randomly initialized parameters. 

\begin{figure*}[htb]
%\settoheight{\includegraphics[width=\linewidth]{example-image-a}}%
\centering\begin{tabular}{@{}c@{ }c@{ }c@{ }}
\textbf{Fully Supervised} & \textbf{Low-Aug UMTRA} & \textbf{Proposed} \\
%\rowname{Exp 1}&
\includegraphics[width=.32\linewidth]{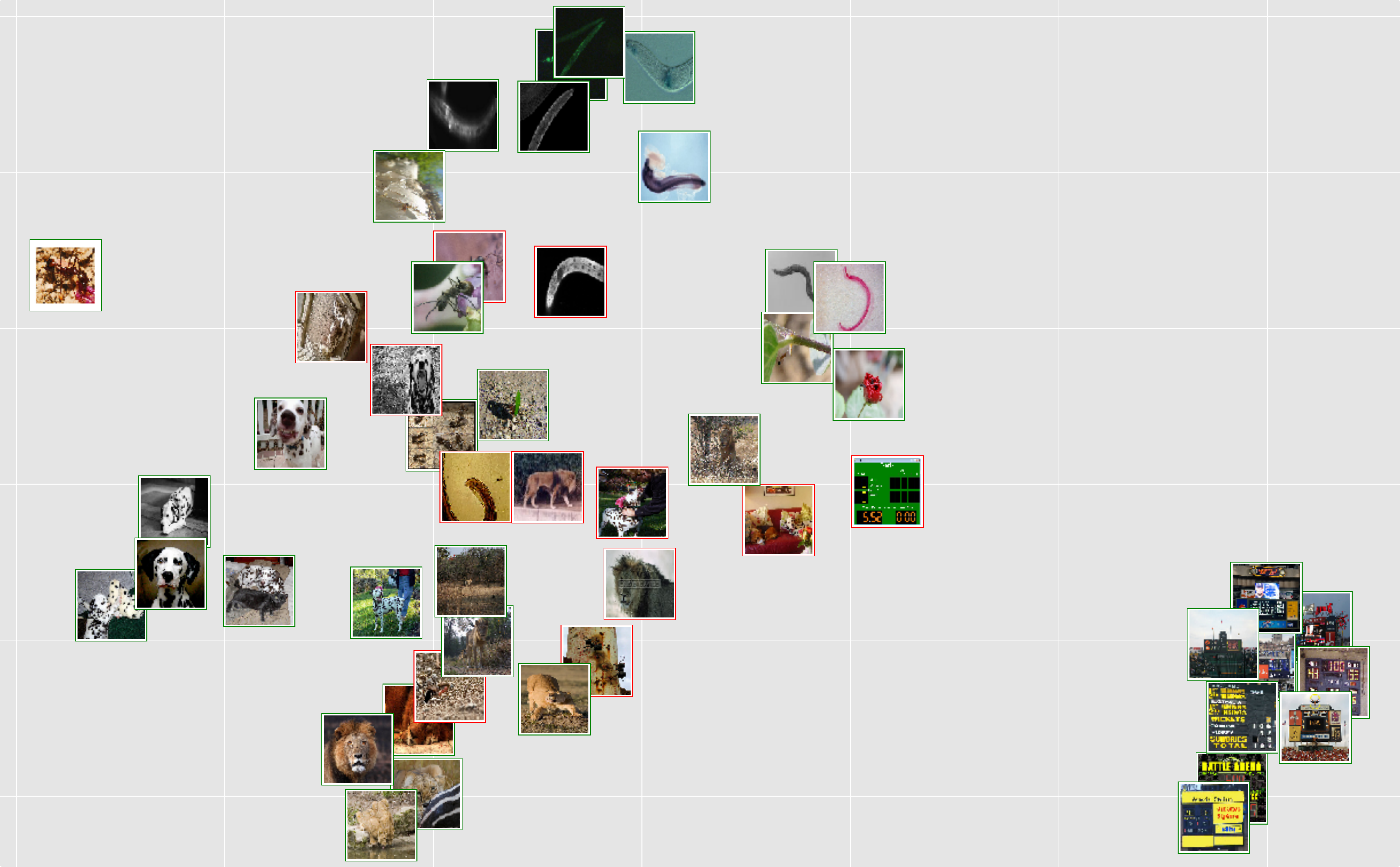}&
\includegraphics[width=.32\linewidth]{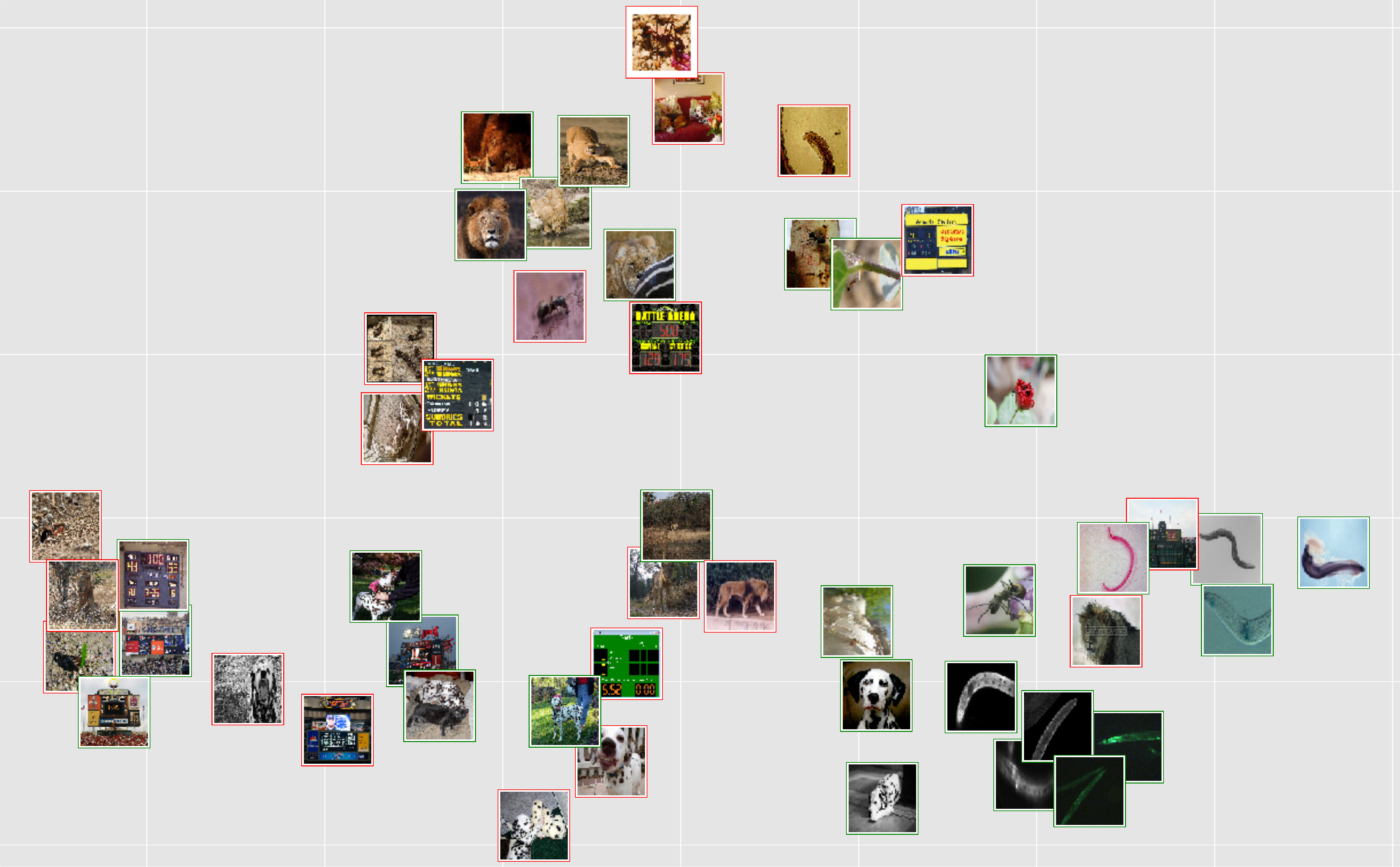}&
\includegraphics[width=.32\linewidth]{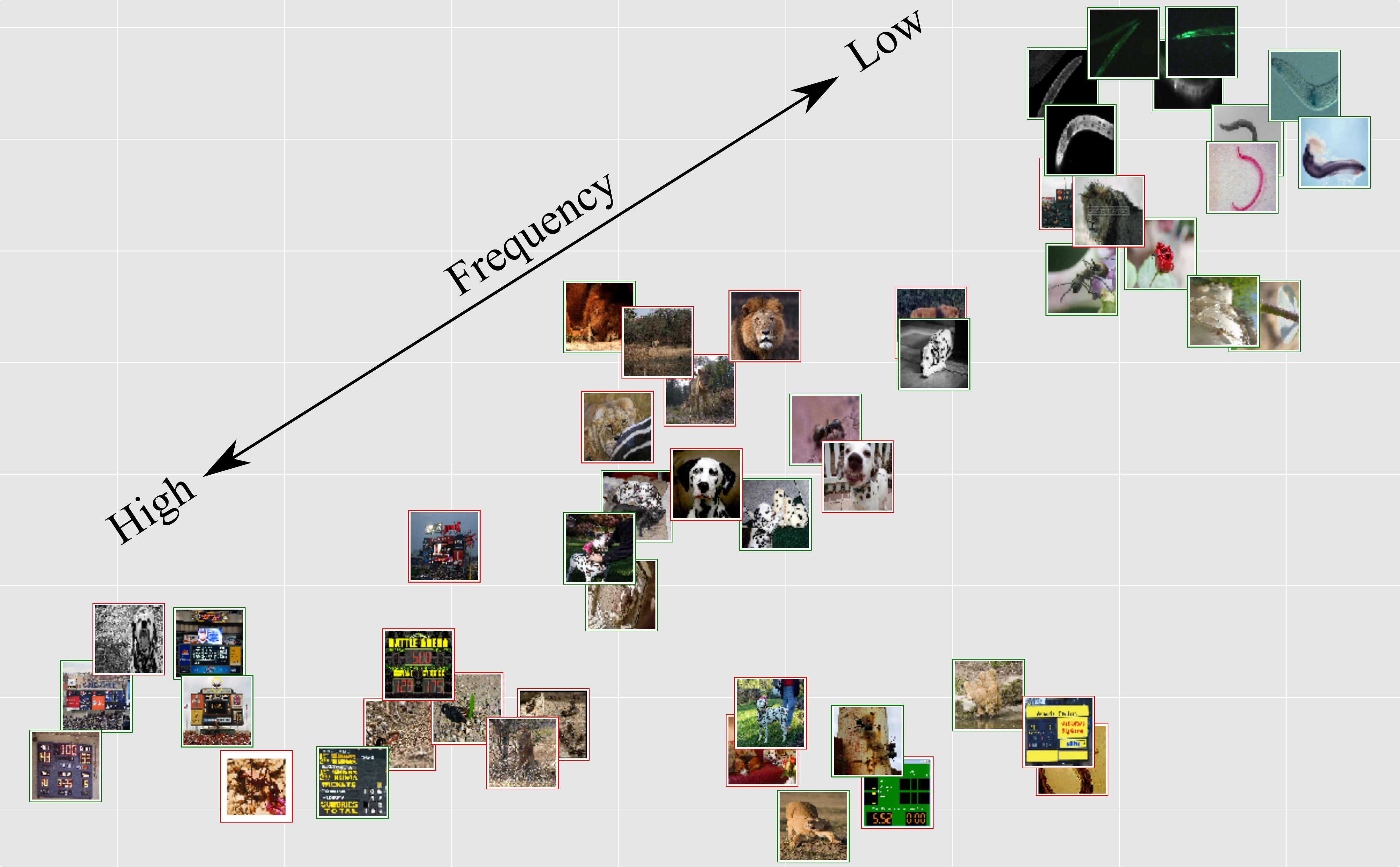}\\[-1ex]
(a) & (b) & (c)\\
%\rowname{Exp 2}&
\includegraphics[width=.32\linewidth]{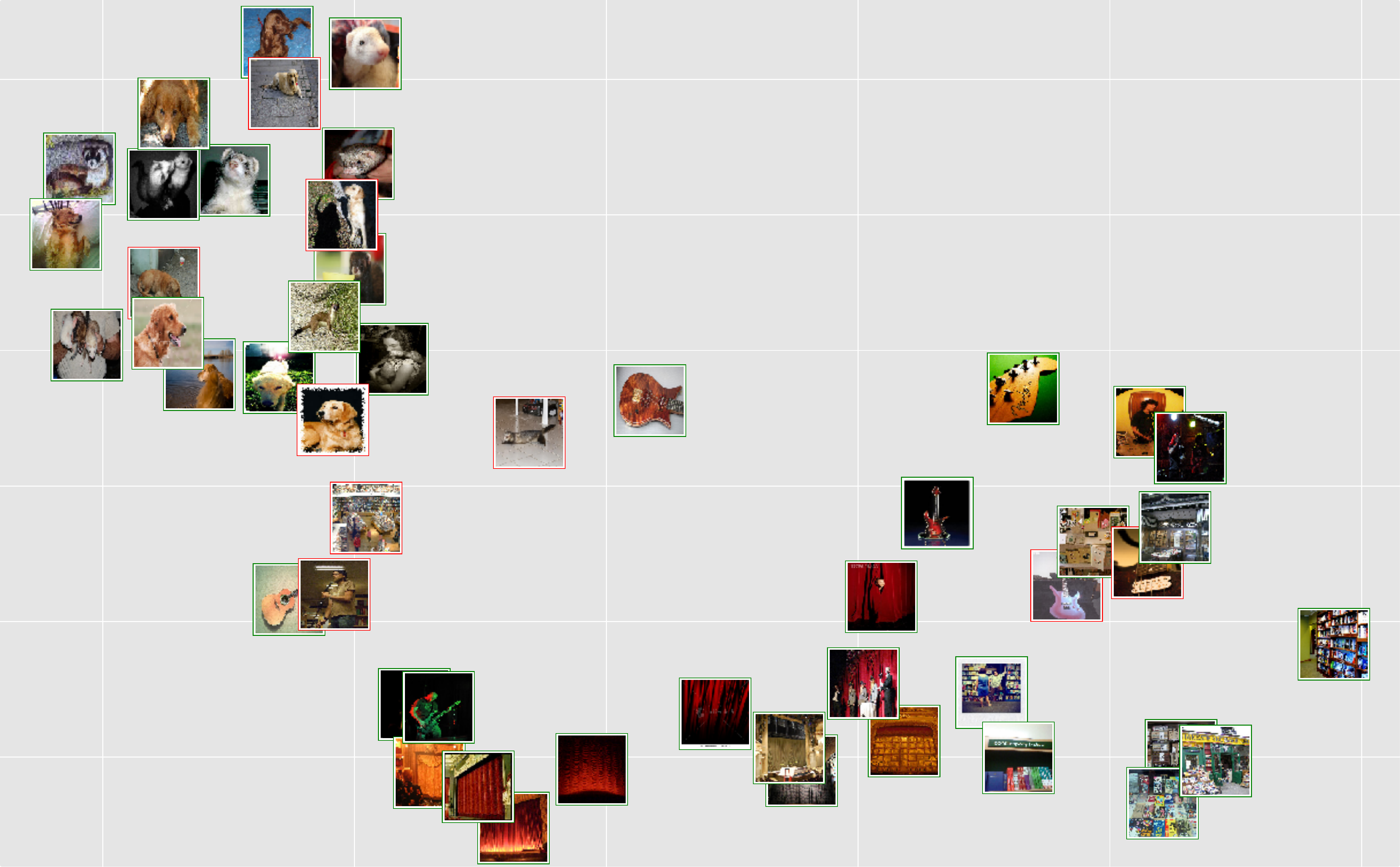}&
\includegraphics[width=.32\linewidth]{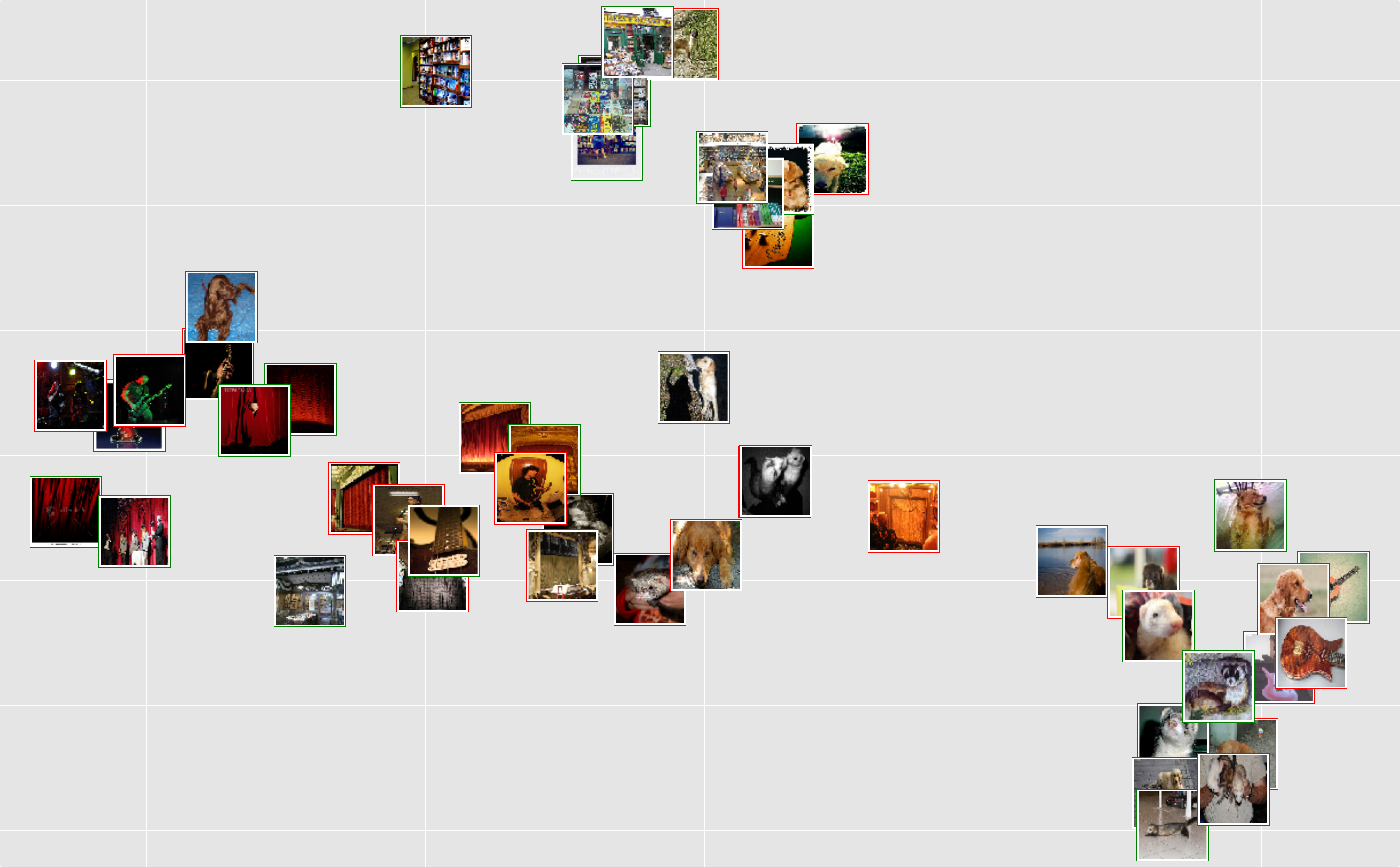}&
\includegraphics[width=.32\linewidth]{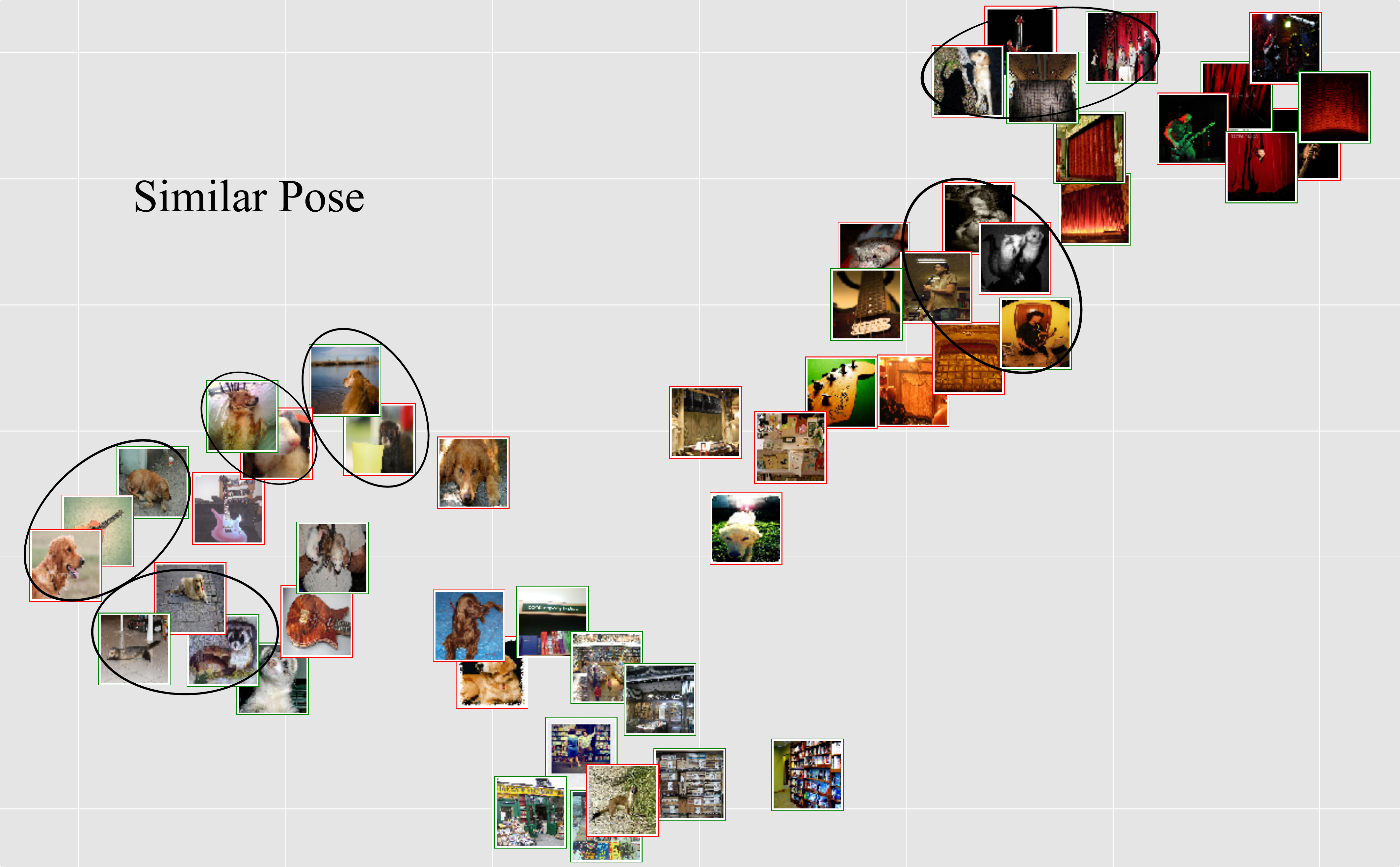}\\[-1ex]
(d) & (e) & (f)\\
%\rowname{Exp 3}&
\includegraphics[width=.32\linewidth]{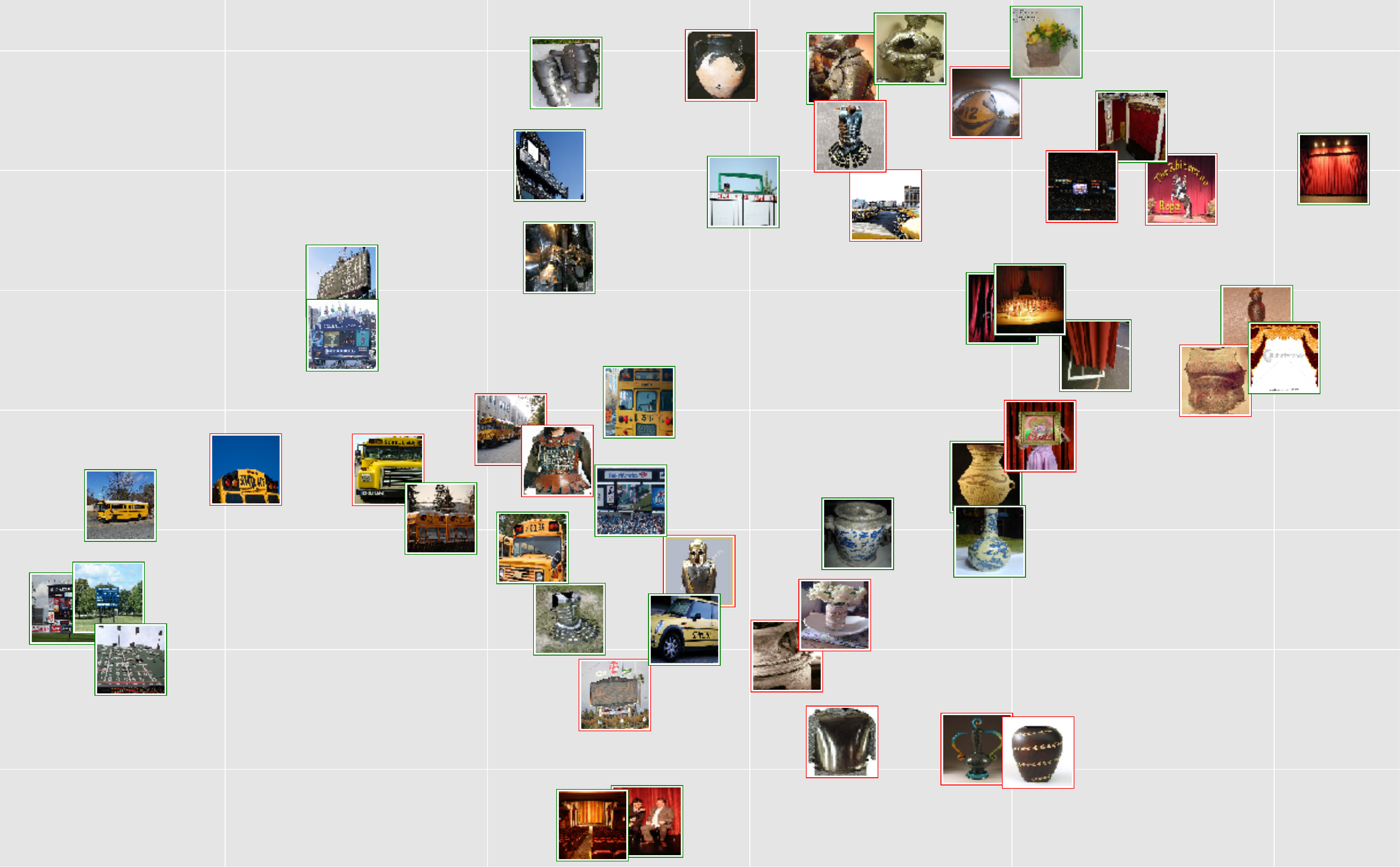}&
\includegraphics[width=.32\linewidth]{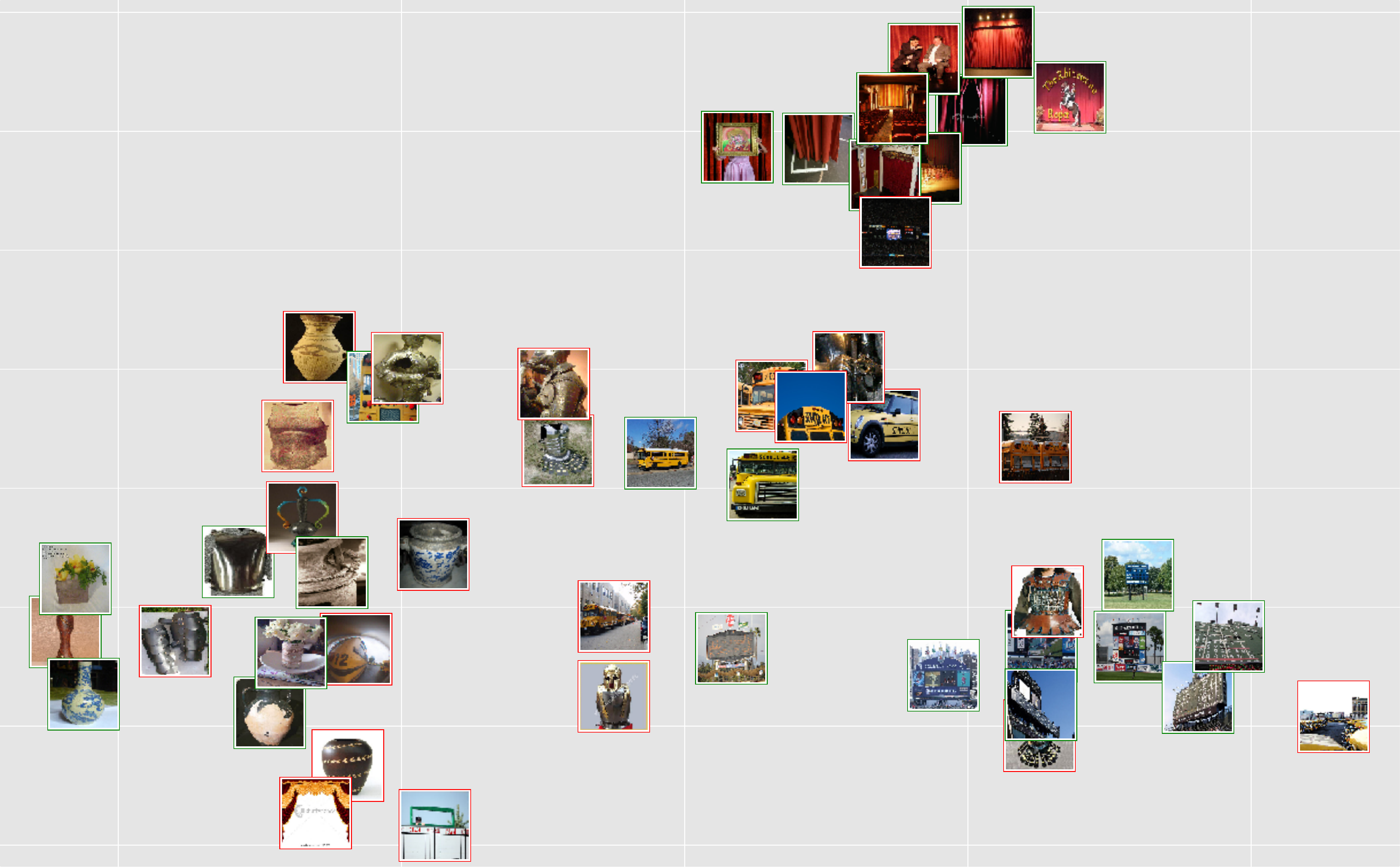}&
\includegraphics[width=.32\linewidth]{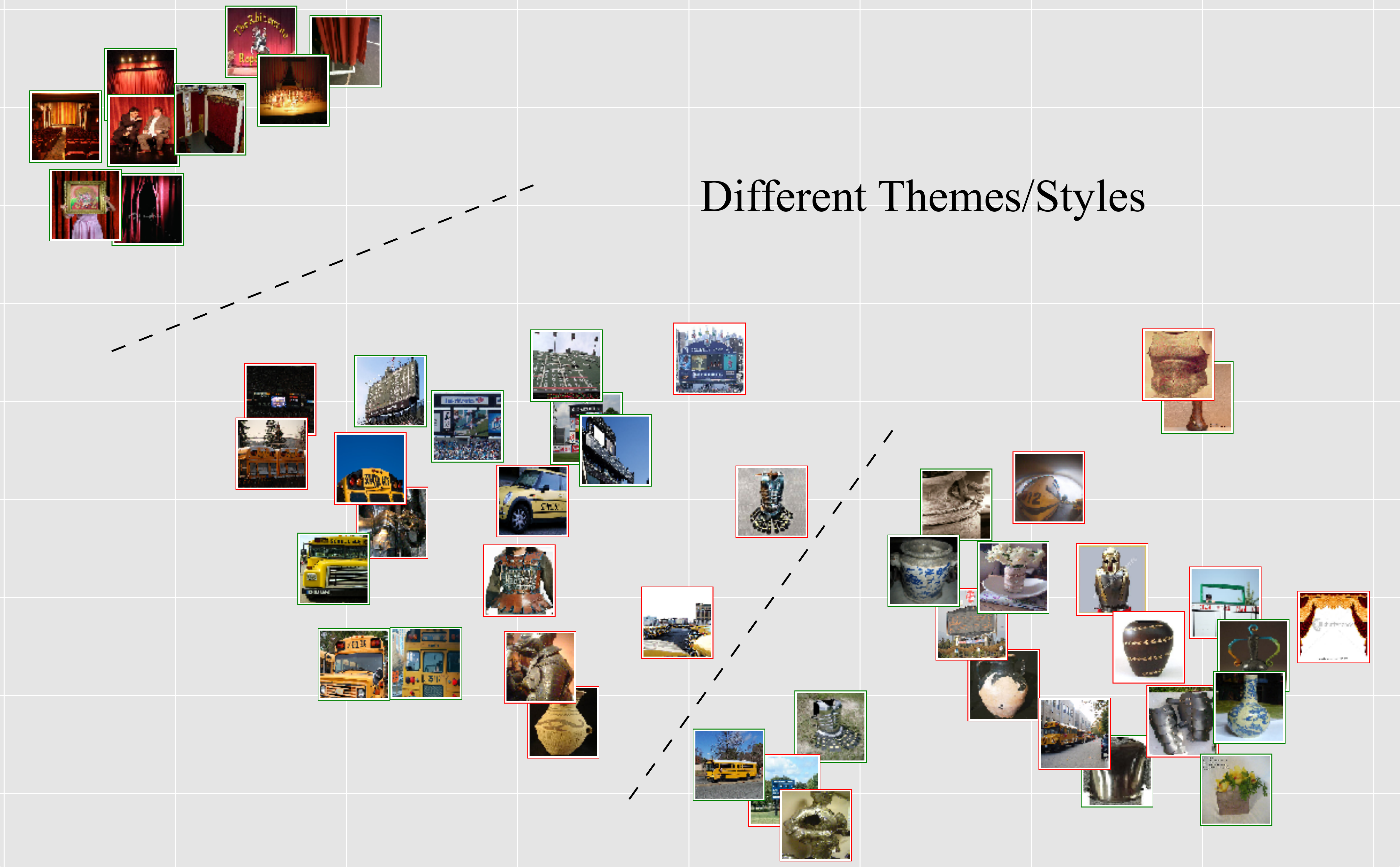}\\[-1ex]
(g) & (h) & (i)\\
\end{tabular}
\caption{{Example t-SNE plots using the logit outputs for 10 test images per class after fine-tuning on $N=5, K=5$ examples. Each row is a different task. Green/red borders indicate whether images were correctly/incorrectly classified. The fully supervised approach is able to group images by class characteristics that are invariant to transformations. The proposed divergent approach appears to rely on more generic features such as spatial frequency, pose, or image style. Best viewed digitally.}}%
\label{figures:tsne}
\end{figure*}

A closer examination of the classification behaviors requires inspection of individual tasks. Figure~\ref{figures:tsne} displays several example tasks using t-distributed stochastic neighbor embedding (t-SNE) plots~\cite{maaten2008visualizing,policar2019opentsne}. In each case, the model is fine-tuned on $K=5$ examples from $N=5$ classes. Then 10 test images are sampled from each class and the logit outputs are visualized using t-SNE. Although t-SNE is not deterministic, the resulting plots provide a \textit{potential} interpretation of the model's sorting criteria. Red and green borders around each image indicate the classification outcome (incorrect/correct respectively). In some instances, classification outcome is poor, but the images are still embedded in meaningful ways.

We also perform an ablation study (Table~\ref{tab:Ablation}). First, the quality aspect is removed by accepting every behavior found by the divergent search in the inner loop (`Divergent Only'). Then, the divergent search is replaced by random search. In random search, labels are sampled dynamically from a discrete uniform distribution across all possible classes. Note that this is different from the \textit{fixed} random labels setting where an image is assigned to a fixed group for the entire pre-training process. Each component of the proposed method increases the evaluation accuracy, especially the divergent search used in the inner loop. 

%An ablation study is also helpful to quantify the contribution of each component. To measure the value of divergent search, it is often compared with random search. In this case, random search is simulated by generating random labels for each sampled image on-the-fly. This is different from the \textit{fixed} random labels approach where images are assigned a random class-membership which remains constant throughout pre-training. Sampling labels dynamically from a discrete uniform distribution (over all possible class labels) creates more diverse tasks. However it does not consider which tasks the model is already capable of performing. As such, it does not push the model to perform diverging tasks, but simply searches in random directions for each batch. Table~\ref{tab:Ablation} indicates that random search, although diverse, does not lead to 

\begin{table}[]
\caption{Ablation study on Mini-ImageNet with $N=5,K=5$ evaluation tasks.}
\begin{tabular}{lll}
\textbf{Ablation Study}\\ \hline
 Algorithm & Accuracy \\
%& Randomly Search &      & Diversity Only &      & Proposed &   \\ 
%          & train                     & test  & train          & test & train    & test  \\ \hline
 & train & test \\ \hline
Random Search                         & 32.54              & 26.63 \\
Random Initialization                 & 39.93              & 36.52 \\
\textit{Proposed} (Diversity Only)    & 46.52              & 42.41 \\
\textit{Proposed} (Quality Diversity) & 47.58              & 43.41 \\
\end{tabular}
\label{tab:Ablation}
\end{table}
%Proposed              & 46.12              & 42.08 \\

\section{Discussion}
Evaluation on class-based tasks indicates the usefulness of a model's learned features. Supervised tasks and tasks that are constructed to \textit{approximate} supervised tasks lead to the highest performance. Divergent search is able to learn features which are more beneficial than harmful. However, as expected, its performance is still significantly lower than methods with a strong bias toward evaluation tasks. The image sampling protocol (Section~\ref{sec:sampling}) used in the proposed method produces batches of images that do not neatly fit into five classes. Nonetheless, the proposed algorithm tries to find a new, high quality behavior for each batch. In contrast, UMTRA relies on statistically-favorable sampling and CACTUs-MAML draws samples from distinct regions in an embedding space, all in an effort to approximate supervised tasks. These results demonstrate that \textit{despite} unfavorable conditions, an undirected search in a high-dimensional behavior space can still be fruitful. In comparison, many of the learned embeddings (BiGAN, ACAI, DeepCluster) perform worse than divergent search, sometimes even worse than random initialization. These learned embeddings typically find features that are relevant to the data; but without a bias toward evaluation tasks, their efforts can be detrimental.

%The image samples given to divergent search (using the sampling protocol detailed in section~\ref{sec:sampling}), it is actually difficult to construct tasks that are aligned with the evaluation tasks. 
%the divergent search is actually predisposed to learn tasks that are not aligned with evaluation tasks. In contrast, UMTRA and CACTUs-MAML construct tasks that are likely to contain distinct groups of images with that have some inherent similarity. 
%In UMTRA, 5 images are sampled which are likely to be distinct; additional samples are inherently related because they are transformed versions of the originals. For CACTUs-MAML, 5 groups are selected based on clusters in a learned representation. Assuming that nearby embeddings have some commonality, samples within a group will be more closely related to each other than samples from another group. High inter-class variance and low intra-class variance can provide a strong training signal  

%This is surprising because it does not have any bias 
%Surprisingly many of the representations (BiGAN, ACAI, DeepCluster) are not always useful for the evaluation tasks. In some cases they actually perform worse than random initialization. Although they are unsupervised approaches, it is possible that they have overfit to features in the pre-training classes which are not present in the evaluation tasks. Training CACTUs-MAML on tasks derived from these representations helps the model to generalize better to evaluation classes. This suggests that   
%specificity of bias - representations
%sampling assumptions

The random fixed labels approach also leads to a drop in performance (Table~\ref{tab:OmniglotResults} and Table~\ref{tab:MiniImageNetResults}). This suggests that the model is specializing in a behavior that is counterproductive. 
%(In fact, learning from random labels can lead to performance that is even worse than not training at all\cite{hsu2018unsupervised}, i.e. Xavier initialization\cite{glorot2010understanding})
Hsu et al. found a similar reduction in performance when training on random labels~\cite{hsu2018unsupervised}. These random tasks often include a) multiple classes under the same label and b) the same class under multiple labels. As such, it is reasonable that the bias learned from this pre-training is harmful when performing class-based evaluation tasks. However, the tasks constructed in a divergent search are also likely to violate class-boundaries. In spite of this, the divergent search is able to learn features that are more beneficial than harmful. One key consideration is that optimization on \textit{fixed} random labels will lead to specialized discriminative behaviors for \textit{that} distribution of tasks. In contrast, a divergent search tries to explore as many novel discriminative behaviors as possible.

Ideally, a divergent search should accumulate specialists in \textit{many} different behaviors. But in this setting the model size is restricted for fair comparison with benchmarks. It is also impractical to use niche-preserving methods (e.g. novelty search with local competition~\cite{lehman2011evolving}) when dealing with such a high-dimensional behavior space. As such the model is incapable of learning a specialized set of behaviors for every task it encounters as niche-preserving methods do. Instead, the best use of the model capacity may be a set of generic features that have the most overall utility. Figure~\ref{figures:tsne} shows that the proposed method may have learned to distinguish images by their spatial frequency, pose, and style, all of which are fairly generic descriptors. This type of undirected learning may eventually lead to more robust, general-purpose features than those found by tuning a model precisely to human-crafted objectives. However, the wide gap in performance when compared to strongly biased methods suggests that its current state leaves much to be desired.

The fully supervised method achieves higher scores by grouping images based on characteristics associated with class identity. For example, in Figure~\ref{figures:tsne} panel (a), the model groups lions and Dalmatians separately despite their high intra-class variance. Low-augmentation UMTRA, panel (b), is also able to make this distinction. Both have learned some amount of invariance to identity-preserving transformations thanks to the bias in their pre-training strategies.

Divergent search is also often compared with random search. In this case, the tasks for random search are constructed by dynamically sampling labels from a discrete uniform distribution over all possible classes. Sampling random tasks in this way effectively asks the model to perform diverse random behaviors. Uniform random search in behavior space has recently been proposed as an interpretation of the overall outcome of novelty search~\cite{doncieux2019novelty}. Given these intuitions, random search might be expected to perform just as well as a divergent search. However, Table~\ref{tab:Ablation} indicates that this formulation of random search actually has a deleterious effect on evaluation performance. The main difference between a random behavior task and a divergent task is that the divergent task is strongly connected to the model's current capabilities. In our case, the divergent task originates from a task that the model can already perform. Then it is optimized away from the space of behaviors that the model can achieve given its current features. This optimization process is stochastic, heuristic and imperfect. The resulting task may lie somewhere between easy (the model's current behavior) and hard (a behavior that is impossible with the model's current features). This may unintentionally regularize the difficulty of divergent tasks and form a sort of curriculum~\cite{bengio2009curriculum}. As such, these divergent tasks may have more structure than the random search in high-dimensional behavior space, allowing for more constructive learning.

Our experiments explore the application of divergent search to image classification. Although the motivation is similar to many existing works on divergent search, there are several key differences. One difference is in the nature of the task. In maze solving tasks, which is a common problem setting for divergent and open ended evolution studies~\cite{lehman2011abandoning,brant2017minimal}, the complexity of a maze is fairly simple to measure and provides a meaningful metric of progression. This is useful for creating a trivial starting point (e.g. empty maze) for behaviors to develop. It also inherently provides a direction for learning which can gradually increase in complexity, e.g. sequentially adding more barriers. Neither of these are available in the image classification setting. A trivial behavior could be classifying all images as the same class, but this is actually a degenerate behavior as compared to an initialization from random~\cite{glorot2010understanding}  parameters. Furthermore, task complexity is not easily measurable and more complex tasks do not lead to better or more useful behaviors. Another difference is that the behavior space is very high-dimensional compared to the environments used in many reinforcement learning problems. In agent-based frameworks, the behaviors can often be conflated onto a low-dimensional space where nearly all regions are meaningful~\cite{lehman2011evolving}. Exploring this behavior space is very likely to be fruitful. In the case of image classification, the behavior space is vast and many regions of the space do not correspond to anything meaningful. Although this is a challenging setting, the behavior space found in nature is even larger. Bringing divergent search into these larger playgrounds can provide insight into the needs that are currently unmet.

%boot strapping
%sampling is inherently unsolvable - as opposed to restricted maze conditions

%Random labels in analogy to MAPElites is like finding one specific solution. Very unlikely that it will be useful for our target/evaluation tasks.
\section{Conclusion}
Divergent search is a major component of QD methods which aim to learn a repertoire of behaviors in the absence of specific behavioral objectives. The reasoning behind this approach is that a system equipped with a whole host of skills will be able to handle novel tasks better than a system which is tuned for one specific task. Meta-learning, i.e. learning how to best learn a new task, has a very similar end goal. However it is typically approached from the angle of generalizing from tasks that are similar to the target task. In this work we investigate the use of divergent search in a few-shot image classification setting. Comparing divergent search to meta-learning approaches, which are biased toward the target task, reveals a significant gap in performance. But compared to weaker biases or methods which are not specifically biased, divergent search is often comparable or better. These results demonstrate that divergent search is a viable approach, even in a high-dimensional behavior space. There is also considerable room for improvement, particularly in the measurement of behavior quality. In future work, better quality metrics could help condense the behavior space to dimensions with more utility. A more modular architecture could also preserve archived behaviors in a better way, facilitating exaptation.

\begin{acks}
JT is funded by the IC President's Scholarship.
\end{acks}

\bibliographystyle{ACM-Reference-Format}
%\bibliography{ref} 
%%% -*-BibTeX-*-
%%% Do NOT edit. File created by BibTeX with style
%%% ACM-Reference-Format-Journals [18-Jan-2012].

\end{document}